# Robust Iris Centre Localisation for Assistive Eye-Gaze Tracking

Nipun Sandamal, Stefania Cristina, Kenneth P. Camilleri
Department of Systems and Control Engineering, Faculty of Engineering, University of Malta, Msida, Malta
*Contact Author: nipun.ranasekara@um.edu.mt*

## Abstract

In this research work, we address the problem of robust iris centre localization in unconstrained conditions as a core component of our eye-gaze tracking platform. We investigate the application of U-Net variants for segmentation-based and regression-based approaches to improve our iris centre localisation, which was previously based on Bayes' classification. The achieved results are comparable to or better than the state-of-the-art, offering a drastic improvement over those achieved by the Bayes' classifier, and without sacrificing the real-time performance of our eye-gaze tracking platform.

## Introduction

Video-based eye-gaze tracking technology offers an assistive solution to individuals with limited mobility by enabling them to interact with a device, such as a computer, through their eye and head movements alone. Fuelled by an interest in tracking the eye-gaze under unconstrained real-life conditions, years of research work have led us to develop a real-time passive eye-gaze tracking platform that estimates the eye-gaze by means of a geometrical model, which integrates information about the image position of the iris centre and the user's head pose (Cristina and Camilleri, 2016). In our work, iris centre localisation was carried out by first segmenting the iris region using a Bayes' classifier and subsequently finding the centroid of the resulting binary blob (Cristina and Camilleri, 2014). This approach worked robustly when the iris region was sufficiently illuminated and visible, but was challenged by real-life factors such as heavy eyelid occlusion and shadow, which often led the Bayes' classifier to incorrectly segment the iris region and consequently displaced the iris centre estimation.

Recent advancements in Convolutional Neural Networks (CNNs) led to such models being increasingly applied to the problem of iris segmentation over the years (Arsalan et al., 2017; Wang et al., 2020; Chen et al., 2022), with increasing complexity and computational cost. For instance, Arsalan et al. (2017) proposed a two-stage approach, which performs image enhancement and noise removal operations in the first stage, and then feeds the pre-processed image to a VGG-face model for pixel-level classification. More recently Chen et al. (2022) proposed a Dual Attention Densely Connected Network that incorporates two attention modules and skip connections.

Another interesting approach formulates iris/pupil centre localisation as a heatmap regression problem, where a heatmap is generated for any given image and the iris centre coordinates are found as the 'hottest' point in the heatmap (Xia et al., 2019; Choi et al., 2019). The model of Xia et al. (2019), for instance, is based on a VGG-16 architecture, but their performance was found to be susceptible to eye region occlusions, reflections, and shadow. Choi et al. (2019), on the other hand, made use of a U-Net model with an added skip connection between the encoder and decoder.

In light of the challenges for eye-gaze tracking posed by real-life conditions, the research work presented in this paper focuses on investigating segmentation-based and regression-based deep learning models for iris centre estimation. Our aim is to improve the accuracy of iris centre localisation, as a core component of our eye-gaze tracking platform, keeping the inference time to a minimum to retain the platform's real-time capabilities and simplifying the processing pipeline.

# Method

Following preliminary experiments with several models, we have identified the U-Net family of models to be providing the most promising results for both segmentation and heatmap regression. Hence, the research work that we present in this section is based upon variants of the U-Net model.

Iris Centre Localisation via Segmentation

The U2-Net model features a two-level nested 'U' structure comprising encoder, decoder and bridge components, with symmetric encoder and decoder blocks (Qin et al., 2020). Unlike other segmentation models, the U2-Net model does not rely on a pre-trained backbone to abstract local and global features. This reduces the computational cost, making this model more suitable for real-time applications such as ours. In training the U2-Net model, we have tailored the loss function to combine region-based and boundary-aware loss functions (Terven et al., 2023), as follows:

$$\text{Total Loss} = \text{Region-Based Loss} + \text{Boundary-Aware Loss} \quad (1)$$

The Dice Loss function was used as the region-based loss function, which measures the similarity of the ground truth and predicted segmentation masks, with the aim of training the model to produce spatially coherent segmentation results. This is combined with the Boundary Loss function to emphasise accurate segmentation of the iris boundary. Following segmentation, the iris centre was localised by finding the centroid of the resulting binary blob, as in our approach in Cristina and Camilleri (2014).

Iris Centre Localisation via Heatmap Regression

A U-Net model (Ronneberger et al., 2015) was trained to generate a heatmap for a cropped eye region image, in which the iris centre was then localised as the 'hottest' point in the heatmap. The vanilla U-Net model was modified slightly by introducing CoorConv operations in its initial convolutional layers of its encoder block, with the aim of encouraging the model to learn spatial relationships between pixels. During training, a Mean Squared Loss function was used. Furthermore, the ground truth heatmaps were generated by centering a Gaussian kernel on the ground truth iris coordinates, as explained by Xia et al. (2019).

# Results and Discussion

The U2-Net segmentation model was trained and tested on the MICHE 2 dataset (De Marsico et al., 2017), this being a challenging dataset featuring various eye rotations under strong shadow and reflections, and comprising ground truth segmentation masks. Table I tabulates the achieved segmentation results as *Ours (Seg)*, and compares them to the results of several state-of-the-art methods. These state-of-the-art methods have been tested on the MICHE 1 dataset (De Marsico et al., 2015), which is however encompassed and extended in the MICHE 2. Hence, for this reason the MICHE 2 is considered to be the most challenging of the two datasets. Furthermore, since the MICHE 2 dataset does not include the iris centre coordinates, these were found via centroid detection on the ground truth and predicted segmentation masks. This permitted us to compute the iris centre localisation error of the segmented masks, as tabulated in Table II by *Ours (Seg) + Centroid*, to be able to draw comparisons with the results of the U-Net.

Table I: Comparison of iris segmentation results.

**Iris Segmentation**

| Method | E1 | Precision | Recall | F1-Score |
|---|---|---|---|---|
| MICHE 1 DATASET | | | | |
| Arsalan et al. (2017) | 0.0035 | - | - | - |
| Wang et al. (2020) | 0.82 | - | - | - |
| Chen et al. (2022) | - | - | - | 0.9319 |
| MICHE 2 DATASET | | | | |
| **Ours (Seg)** | 0.00627 | 0.9499 | 0.9533 | 0.9526 |

Table II: Comparison of iris centre localisation results.

**Iris Centre Localisation**

| BioID DATASET | $d \leq 0.25$ | $d \leq 0.10$ | $d \leq 0.05$ |
|---|---|---|---|
| Bayes' + Centroid | 85.71% | 57.72% | 35.27% |
| Xia et al. (2019) | 100% | 99.90% | 94.40% |
| Cai et al. (2019) | - | - | 92.80% |
| Choi et al. (2019) | - | - | 93.30% |
| Lee et al. (2020) | 100% | 98.95% | 96.71% |
| **Ours (Seg) + Centroid** | 99.12% | 93.29% | 52.18% |
| **Ours (Reg)** | 100% | 99.41% | 96.33% |
| GI4E DATASET | | | |
| Bayes' + Centroid | 89.91% | 35.96% | 15.35% |
| Xia et al. (2019) | 100% | 100% | 99.10% |
| Cai et al. (2019) | - | - | 99.50% |
| Choi et al. (2019) | - | - | 99.60% |
| Lee et al. (2020) | 100% | 99.84% | 99.84% |
| **Ours (Seg) + Centroid** | 97.36% | 97.36% | 96.92% |
| **Ours (Reg)** | 100% | 99.62% | 99.21% |
| I2Head DATASET | | | |
| Bayes' + Centroid | 92.02% | 59.92% | 26.34% |
| **Ours (Seg) + Centroid** | 99.44% | 99.44% | 98.51% |
| **Ours (Reg)** | 100% | 99.62% | 98.71% |
| MPIIGaze DATASET | | | |
| Bayes' + Centroid | 93.75% | 59.38% | 37.5% |
| **Ours (Seg) + Centroid** | 100% | 99.21% | 97.65% |
| **Ours (Reg)** | 100% | 100% | 98.23% |

The U-Net model for heatmap regression was, on the other hand, trained and tested on the BioID (Jesorsky et al., 2001), GI4E (Ponz et al., 2012), I2Head (Martinikorena et al., 2018) and MPIIGaze (Zhang et al., 2015) datasets, since these included the required iris centre coordinates as ground truth information. These datasets yielded a total of 10,850 images, which were partitioned into training, validation, and testing sets with proportions of 75%, 10%, and 15%, respectively. Table II tabulates the results achieved by the trained U-Net model as *Ours (Reg)*, along with those reported by several state-of-the-art methods which regress the iris centre coordinates directly at the output. In the literature, most evaluations are carried out on the BioID and GI4E datasets, to which we have also included the I2Head and MPIIGaze to test the generalisability of our model. Our earlier approach employing Bayes' classification followed by centroid detection (Cristina and Camilleri, 2014) was also tested for comparison. The Maximum Normalised Error, denoted by d, is used as the evaluation metric, where the maximum error between left and right eye predictions is normalised by the distance between the eye centres (Cai et al., 2019). When d ≤ 0.05, this indicates that the error is within the pupil diameter. If d ≤ 0.10, the error is within the iris diameter, whereas if d ≤ 0.25, the error is within the distance between the eye corners.

The segmentation results in Table I indicate that the trained U2-Net model performs well with respect to the state-of-the-art, albeit on an extended and more challenging dataset. This is corroborated by the qualitative results in Figure 1, showing challenging examples from the MICHE 2 dataset comprising contrasting variations of light and reflections.

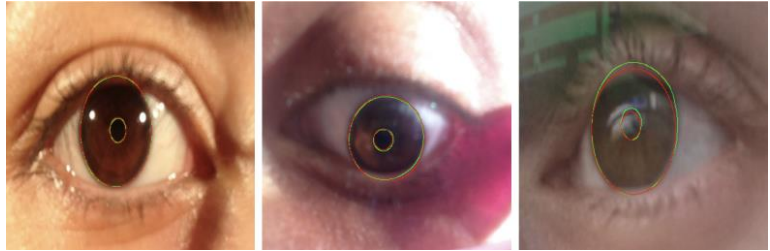

Figure 1: Segmentation showing ground truth (green) and predicted (red) iris boundaries.

The regression results in Table II indicate that the approach based on Bayes' classification performed very poorly. This was expected since the datasets under consideration contain iris occlusions and illumination variations that proved to be too challenging to the Bayes' classifier. The U-Net model trained for heatmap regression, on the other hand, can be seen to perform at par or better than the state-of-the-art. Its architecture is also sufficiently lightweight permitting it to process a cropped eye region image in an average of 20ms on an Intel Core i7-10750H machine with an NVIDIA GeForce GTX 1660 Ti GPU. It may also be noticed that the approach of performing centroid detection on the U2-Net segmentation results (*Ours (Seg) + Centroid*) also performed relatively well when considering that the U2-Net model was only trained on the MICHE 2 dataset. The error in this case is possibly stemming from the fact that the centroid does not necessarily represent the iris centre well given the convexity of the eyeball surface.

# Conclusion

In this work, we have investigated the application of U-Net variants for segmentation-based and regression-based approaches to iris centre localisation. The results indicate that the heatmap regression U-Net, albeit having a relatively simpler processing pipeline, achieves comparable or better results than the state-of-the-art without sacrificing real-time performance. Future research work will be looking at integrating the heatmap regression U-Net into our real-time eye-gaze tracking platform, and testing this out within an assistive context.

# Acknowledgments

This work is part of the LuminEye project, financed by the Malta Council for Science & Technology, for and on behalf of the Foundation for Science and Technology, through the FUSION: R&I Research Excellence Programme 2022.